\documentclass{article}
\usepackage{iclr2027_conference,times}
\usepackage[utf8]{inputenc}
\usepackage[T1]{fontenc}
\usepackage{microtype}
\usepackage{graphicx}
\usepackage{booktabs}
\usepackage{amsmath,amssymb,mathtools}
\usepackage{xcolor,colortbl}
\usepackage{placeins,float}
\usepackage{hyperref,url}
\definecolor{softteal}{HTML}{E9F5F3}
\hypersetup{colorlinks=true,allcolors=blue,pdfauthor={Jinhao Chen, Benlei Cui, Ruijian Jia, Ziheng Wang, Tianyu Wo, Pengfei Sun, Longtao Huang, Hui Xue, Yitong Yang, Haiwen Hong},pdftitle={Grounding with Confidence}}

\title{Grounding with Confidence: Controllable Generative Video Temporal Grounding}
\author{Jinhao Chen$^{1,2,*,\ddagger}$ \quad Benlei Cui$^{1,*,\dagger}$ \quad Ruijian Jia$^{1}$ \quad Ziheng Wang$^{1,3}$ \\ \bfseries Tianyu Wo$^{2}$ \quad Pengfei Sun$^{1}$ \quad Longtao Huang$^{1}$ \\ \bfseries Hui Xue$^{1}$ \quad Yitong Yang$^{1}$ \quad Haiwen Hong$^{1,\dagger}$ \\[4pt] \normalfont\small $^{1}$Alibaba Group \quad $^{2}$Beihang University \quad $^{3}$Fudan University}
\iclrfinalcopy \newlength{\yuvionheadextra}\fancypagestyle{yuvionheader}{\fancyhf{}\fancyhead[L]{\raisebox{0pt}{\includegraphics[height=17pt]{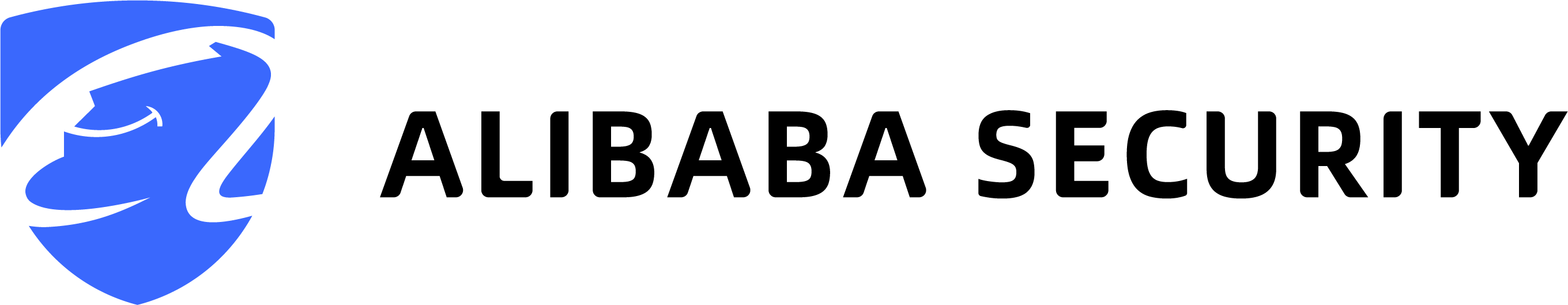}}}\fancyhead[C]{\small\normalfont\textcolor{gray}{Preprint}}\fancyhead[R]{\raisebox{0pt}{\includegraphics[height=17pt]{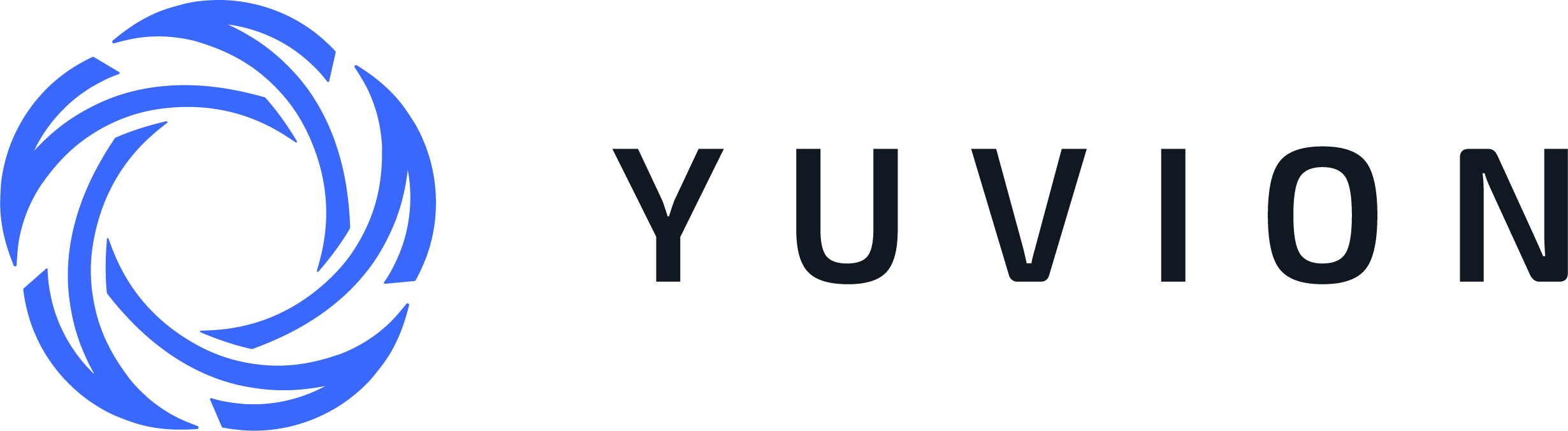}}}\fancyfoot[C]{\thepage}\renewcommand{\headrulewidth}{0.8pt}\renewcommand{\footrulewidth}{0pt}}

\makeatletter\renewcommand{\@maketitle}{\vbox{\hsize\textwidth\centering{\LARGE\sc\@title\par}\begin{tabular}[t]{c}\bf\rule{\z@}{24pt}\@author\end{tabular}\vskip 0.3in minus 0.1in}}\makeatother
\begin{document}
\maketitle\pagestyle{yuvionheader}\thispagestyle{yuvionheader}\begingroup\renewcommand{\thefootnote}{\fnsymbol{footnote}}\footnotetext[1]{Equal contribution.}\footnotetext[2]{Corresponding authors: Benlei Cui (\href{mailto:cuibenlei.cbl@alibaba-inc.com}{\texttt{cuibenlei.cbl@alibaba-inc.com}}); Haiwen Hong (\href{mailto:honghaiwen.hhw@alibaba-inc.com}{\texttt{honghaiwen.hhw@alibaba-inc.com}}).}\footnotetext[3]{This work was done during Jinhao Chen's internship at Alibaba Group.}\endgroup

\begin{abstract}
Video temporal grounding supports applications such as video search, content
review, and automated editing by localizing events described in natural
language. Yet existing generative models typically output timestamps without
explicit interval-level confidence scores to guide candidate selection. We separate
candidate generation from acceptance by scoring individual intervals within
the original decoding pass. A lightweight confidence head reads pooled decoder
states, providing an explicit score trained for interval selection. Offline verifier scores
supervise the head on fixed candidate sequences, and temporal-overlap labels
adapt it to current rollouts during reinforcement learning. GT-anchored
candidate-pool supervision and set-level optimization train the generator.
The resulting scores support ranking, threshold-based selection, and rejection
without invoking an external verifier at inference. On a fixed
OMTG-Bench candidate pool, confidence raises query-macro Recall@0.5 from
9.95\% to 14.42\% over generation order at a 10\% global return budget, and
from 26.48\% to 31.12\% at a 25\% budget. The continuous scores let downstream
applications adjust return budgets or acceptance thresholds to match their
precision--recall preferences, without regenerating candidate intervals.
\href{https://github.com/Alibaba-VELLDEPTH/Grounding-with-Confidence}{Code is available on GitHub.}
\end{abstract}

\section{Introduction}
\label{sec:introduction}

Video temporal grounding (TVG) localizes events described by a language query
in a video. Discriminative TVG models provide proposal or matching scores
\citep{lei2021qvhighlights,lin2023univtg,moon2023qddetr}.
Generative vision-language models express their predictions as timestamp
sequences, accommodating both individual moments and repeated occurrences
\citep{huang2024vtimellm,ren2024timechat,wang2024groundedvideollm,zhu2026timelens2}.
Yet these responses typically provide no explicit confidence for each interval.
Coordinates identify where to look, but provide no interval-level score
explicitly trained for localization reliability.
Recent decision models such as Jev likewise expose probabilities and
confidence alongside predictions to support decisions about model outputs
\citep{almeida2026jev}.

\begin{figure}[t]
  \centering
  \includegraphics[width=\linewidth]{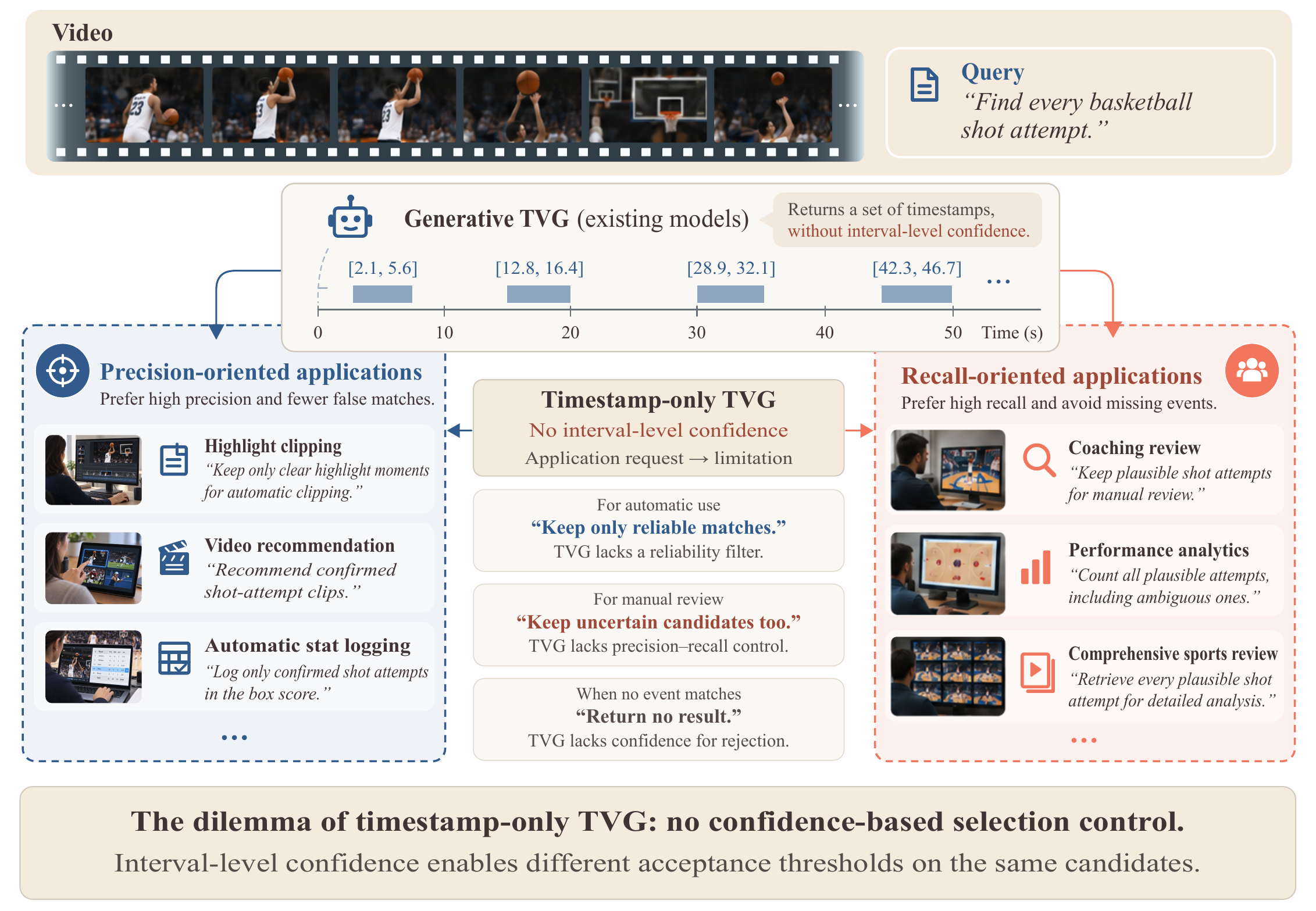}
  \caption{\textbf{Localization alone does not meet downstream selection needs.}
  Applications differ in their tolerance for false and missed matches.
  Timestamp-only generative TVG lacks explicit control for adapting interval
  selection to these requirements. Frames are synthetic and scenarios illustrative.}
  \label{fig:motivation}
\end{figure}

Applications differ in their tolerance for false and missed matches
(Figure~\ref{fig:motivation}). Content moderation, surveillance review, and
comprehensive sports review may favor recall: uncertain segments can be
reviewed, whereas missed events never reach review. Livestream
highlight editing, content recommendations, and targeted video clipping may
favor precision, since false matches produce irrelevant clips or unwanted
recommendations. These requirements call for adjustable acceptance criteria
over the same candidate pool, allowing applications to change which intervals
they return without regenerating timestamps.

Video moment retrieval also requires deciding which candidates to present
first. This ranking matters even when all proposed intervals are retained.
Generation order need not reflect how reliably an interval matches the query,
while sequence likelihood is not explicitly supervised for this purpose.

Queries for absent events expose a further limitation. A model may return
plausible timestamps even when no matching content exists. Negative-query
rejection is a recognized challenge in moment retrieval
\citep{flanagan2025untruth}. This requires deciding whether to return any candidate at all, beyond
choosing which candidate to rank first.

Adapting existing confidence estimators to this setting involves trade-offs.
Token likelihood scores the generated
timestamp sequence, without directly supervising whether its boundaries
localize the queried event. Verbalized confidence adds score tokens to the
response \citep{tian2023calibration}, while sampling-based semantic uncertainty
requires multiple answers \citep{kuhn2023semanticuncertainty}.
Semantic entropy probes avoid repeated sampling, but estimate uncertainty
in text generation rather than temporal localization correctness
\citep{kossen2024semanticentropyprobes}. A separate visual verifier can assess
individual intervals, at the cost of additional model inference on candidate
clips. We seek a score trained for individual temporal predictions and available
from the original grounding pass.

\raggedbottom

We address this gap by pairing each generated interval with \emph{a continuous
confidence score}. A lightweight head reads pooled decoder states for the
interval, reusing features computed during the original autoregressive pass.
Applications can then adjust an acceptance threshold, rank candidates by
reliability, and return an empty result when no candidate passes the threshold.
All three decisions use the same response, without regenerating timestamps
or invoking an external verifier. Candidate generation determines what can
be returned; confidence supplies the acceptance rule.

Training gives the generator and confidence head distinct objectives.
The generator is trained to cover annotated occurrences, whereas confidence
learning also uses imperfect and unsupported intervals. In particular,
unsupported candidates teach the head to reject without becoming part of
the generator's timestamp targets. We therefore use separate supervision
sequences for the two tasks. GT-anchored supervised fine-tuning (SFT) retains
every annotated occurrence and adds a bounded number of supported boundary
alternatives. The confidence head learns from an offline visual verifier's
soft scores on precomputed candidate sequences. Set-level reinforcement
learning (RL) then optimizes candidate coverage and the localization quality
of the confidence-selected set. As the generator's proposals evolve during
RL, binary temporal-overlap labels adapt the head to its current candidates.
Confidence losses update only the head, while the selected-set reward guides
the generator.

Experiments on OMTG-Bench assess fixed-pool selection under equal global
return budgets; additional analyses on Charades, ActivityNet, and QVHighlights
examine single-answer ranking and rejection of synthetic cross-video mismatches.

\begin{samepage}
Our contributions are:
\begin{itemize}
  \item \textbf{Controllable grounding from a single decode.}
  We introduce an interval-level confidence interface for generative TVG
  that separates candidate generation from acceptance. A lightweight head
  scores intervals from the original decoder states, enabling ranking,
  threshold-based selection, and rejection without re-decoding or an
  inference-time verifier.
  \item \textbf{Distinct supervision with set-level optimization.}
  GT-anchored targets train the generator; offline verifier scores and online
  temporal-overlap labels train the confidence head. Set-level RL connects
  the two by rewarding candidate coverage and the quality of the
  confidence-selected set.
  \item \textbf{Selection gains from the same candidates.}
  We show that learning what to return increases the value of a single
  generation. On fixed OMTG-Bench candidates, confidence improves query-macro
  Recall@0.5 most under tight budgets, gaining 4.47 and 4.64 points over
  generation order at 10\% and 25\% global return budgets, respectively.
\end{itemize}
\end{samepage}

\FloatBarrier
\section{Method}
\label{sec:method}

\begin{figure}[!t]
  \centering
  \includegraphics[width=\textwidth]{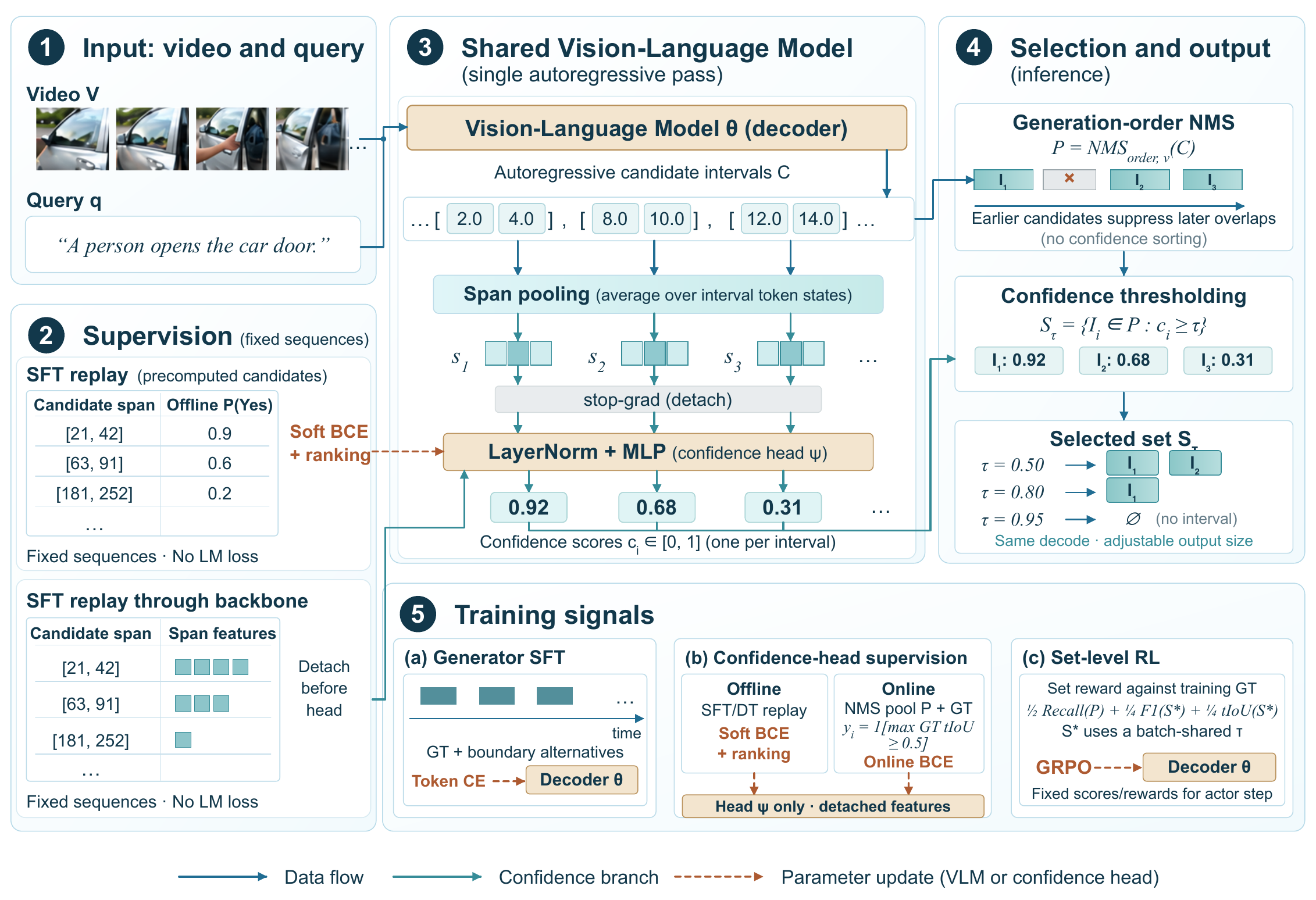}
  \caption{\textbf{Single-pass grounding with confidence.}
  A span-pooled head scores temporal intervals from the same decoding pass.
  Generation-order NMS forms pool $P$; confidence thresholding selects
  $S_\tau$. GT-anchored SFT and set-level RL train the generator;
  offline supervision and online overlap labels train the head. Dashed arrows
  update decoder $\theta$ or head $\psi$; numerical examples are illustrative.}
  \label{fig:method}
\end{figure}

Figure~\ref{fig:method} separates the generation and confidence-supervision paths.
\par
\flushbottom

\subsection{Interval confidence and acceptance}

Given a video $V$ of duration $T$ and a language query $q$, the target is an
interval set $\mathcal{Y}=\{[s_j,e_j]\}_{j=1}^{m}$.  The autoregressive model
$\pi_\theta$ emits a variable-length array
$C=[I_1,\ldots,I_n]$, where $I_i=[\hat s_i,\hat e_i]$ is measured in seconds
and the true cardinality $m$ is not supplied to the model.
A valid candidate satisfies $0\leq\hat s_i<\hat e_i\leq T$.

Let $\mathcal{T}_i$ contain the token positions of the serialized interval
$I_i$, including its coordinates and within-interval delimiters. Let $h_t$
be the final normalized decoder state after consuming token $t$.
We average these states and apply a LayerNorm--MLP readout:
\begin{equation}
  \bar h_i=\frac{1}{|\mathcal{T}_i|}
       \sum_{t\in\mathcal{T}_i}h_t,\qquad
  a_i=g_\psi(\bar h_i),\qquad c_i=\sigma(a_i).
  \label{eq:span-confidence}
\end{equation}
At inference, the head scores intervals from decoder states cached during
timestamp generation, without adding confidence tokens. Teacher-forced replay
is used only for training. Single-pass scoring reuses decoding features;
generating additional candidates still incurs token cost.

We first apply generation-order NMS at overlap threshold $\nu$:
earlier candidates have priority, and a later candidate is suppressed if it
overlaps an already retained one above the NMS threshold.  Denote this pool by
$P=\operatorname{NMS}_{\mathrm{order},\nu}(C)$.  Confidence then selects
\begin{equation}
  S_\tau=\{I_i\in P:c_i\geq\tau\}.
  \label{eq:selected-set}
\end{equation}
NMS does not sort candidates by their confidence. The same completed decode
therefore supports multiple global thresholds without regenerating intervals.
A threshold allows variable output cardinality, including an empty set, rather
than imposing a fixed number of answers per query. The scores can also rank
the same pool under a return budget. These are different uses of one interface:
thresholding fixes an acceptance score, whereas budgeted selection fixes how
many intervals may be returned. Neither rule changes interval boundaries or
recovers candidates removed by NMS.

\subsection{Learning interval confidence}
\label{sec:confidence-learning}

Offline replay supplies visual-support targets over a broad candidate pool;
online overlap labels then adapt the same head to the generator's changing
rollouts. These stages train the acceptance score separately from timestamp
generation.

\paragraph{Offline supervision.}
For each query we use one precomputed candidate sequence, including
supported proposals, unmatched proposals, GT completions, and duration-based
negative candidates.  An offline visual verifier supplies the soft label
$z_i=P_{\mathcal{T}}(\mathrm{Yes}\mid V,q,I_i)$.
These sequences are fixed before SFT and replayed with teacher forcing
to obtain candidate features.  They have zero language-model loss, and their
labels are not transferred to newly synthesized generator alternatives.

The replay's interval states are mean-pooled as in
Eq.~\eqref{eq:span-confidence} and detached before entering the head.
We minimize soft binary cross entropy plus within-query ranking:
\begin{equation}
  \mathcal{L}_{\mathrm{conf}}^{\mathrm{SFT}}
  =\frac{1}{n}\sum_i\operatorname{BCEWithLogits}(a_i,z_i)
   +0.25\mathcal{L}_{\mathrm{pair}},
  \label{eq:head-objective}
\end{equation}
where $\mathcal{L}_{\mathrm{pair}}$ encourages higher logits for candidates
with larger soft labels within a query.
Together with the generator loss $\mathcal{L}_{\mathrm{gen}}$
(Section~\ref{sec:coverage-sft}), SFT preserves gradient separation:
\begin{equation}
  \mathcal{L}_{\mathrm{SFT}}=
    \mathcal{L}_{\mathrm{gen}}+0.25\mathcal{L}_{\mathrm{conf}}^{\mathrm{SFT}},
  \qquad \nabla_\theta\mathcal{L}_{\mathrm{conf}}^{\mathrm{SFT}}=0.
  \label{eq:supervised-objective}
\end{equation}
Generator updates therefore use the GT-anchored generation target, while
head updates use the broader fixed replay distribution. In particular,
unsupported replay candidates teach the head to reject, without teaching the
generator to emit those candidates. The verifier's soft visual-support targets
and the online overlap labels below have different meanings.

\paragraph{Online adaptation.}
The head adapts to current rollouts using binary overlap labels on NMS survivors:
\begin{equation}
\begin{aligned}
  y_i&=\mathbf{1}\!\left[\max_{Y_j\in\mathcal{Y}}
       \operatorname{tIoU}(I_i,Y_j)\geq0.5\right],\\
  \mathcal{L}_{\mathrm{conf}}^{\mathrm{RL}}
    &=\operatorname{mean}_{r}\left[\operatorname{mean}_{i\in P_r}
       \operatorname{BCEWithLogits}(a_i,y_i)\right].
\end{aligned}
  \label{eq:online-head}
\end{equation}
The loss averages candidates within a rollout, then rollouts with valid NMS
candidates. Hidden states are detached, so only $\psi$ receives this loss;
online labels use neither a verifier nor offline replay. A positive label
means that an interval overlaps some GT occurrence sufficiently; it does not
measure the interval's marginal value after other candidates are accepted.
Several candidates can be positive for the same occurrence even though
one-to-one set matching credits that occurrence only once. NMS and the
set-level reward address returned-set quality separately from this local label.

During online adaptation, scores and rewards are computed with the current head and
held fixed for the actor update.  The head then takes one supervised update,
and its new weights score the next rollout batch.  States are reused from the
actor's old-log-probability forward pass.  This ordering prevents a head update
from retroactively changing the current batch's reward. At evaluation both
generator and head are frozen. Confidence influences actor learning because
the current scores determine the selected set used by the reward. Gradient
isolation prevents the supervised confidence loss from directly updating
the backbone.

\subsection{Training the candidate generator}
\label{sec:coverage-sft}
\label{sec:set-optimization}

\paragraph{Generator targets.}
Each generation target retains all GT intervals as anchors and adds a bounded
number of supported boundary alternatives from real proposals or perturbations.
GT anchors precede alternatives and are never dropped to make room for them;
the closing bracket follows the full target. Alternatives vary localization
boundaries rather than introduce independent occurrences. There is no
fixed-count padding or low-IoU filler. The generator loss
$\mathcal{L}_{\mathrm{gen}}$ averages token cross entropy separately within GT,
alternative, and format groups, then weights these groups so their relative
contributions do not depend on token counts.

Coverage and acceptance have different roles. Missing an occurrence during
generation leaves no candidate for any selector to recover, while retaining
several overlapping alternatives can increase raw coverage without improving
returned-set quality. The GT anchors preserve occurrence coverage in the
target, and alternatives expose boundary variation. Generation-order NMS may
still remove an alternative before confidence can compare it with an earlier
proposal; we examine this limit separately from score quality.

\paragraph{Set-level reinforcement learning.}
RL starts from the merged SFT generator and paired head, with full-parameter
actor updates. Each sampled response uses the same NMS and confidence selector
as evaluation. After a recall/localization warm start, the set-level reward is
\begin{equation}
  R(C)=
  \tfrac{1}{2}\operatorname{Recall}_{0.5}(P,\mathcal{Y})
  +\tfrac{1}{4}\operatorname{F1}_{0.5}(S_{\tau_B^*},\mathcal{Y})
  +\tfrac{1}{4}\operatorname{tIoU}(S_{\tau_B^*},\mathcal{Y}).
  \label{eq:set-localization-reward}
\end{equation}
Recall uses maximum-cardinality one-to-one matching at tIoU 0.5, so one
proposal cannot recover two GT occurrences. The first term rewards coverage
before confidence filtering. The remaining terms reward occurrence-level F1
and the temporal intersection over union of the predicted and GT unions.
This training reward's thresholded maximum-cardinality matching differs from
the official evaluator's maximum-total-IoU assignment followed by thresholding
(Appendix~\ref{app:fixed-pool}); the two need not return identical match counts.
One threshold $\tau_B^*$ maximizes mean F1 across the current training batch.
The same selected set supplies both quality terms. No per-query oracle or
test annotation enters training rewards.

We use a variant of group-relative policy optimization (GRPO)~\citep{shao2024deepseekmath}
that centers response rewards by their within-query mean without dividing by
the group standard deviation. The clipped policy objective updates the generator
through the generated response, including coordinate and format tokens.
Confidence selection is deterministic and the head uses the separate
supervised update in Section~\ref{sec:confidence-learning}.

\section{Experiments}
\label{sec:experiments}

\subsection{Experimental setup}

\paragraph{Data and models.}
OMTG-Bench evaluates multi-occurrence grounding on 320 queries, 287 videos,
and 1,173 official GT intervals \citep{xu2026omtg}. We initialize from
TimeLens2-4B \citep{zhu2026timelens2}; SFT uses 60,405 OMTG/TimeLens2 training
units, and RL uses 8,424 OMTG queries with a source-video-disjoint 128-query
internal diagnostic split. Single-interval TimeLens splits supply matched
and synthetic unmatched queries. The confidence head maps 2,560-dimensional
pooled states through 256 hidden units to one logit. Both main OMTG models
use the official visual budget (2 fps, \texttt{min\_pixels}=2048,
\texttt{total\_pixels}=8,388,608), greedy decoding with 512 output tokens and
32 candidates, and generation-order NMS at tIoU 0.3.
Appendix~\ref{app:implementation} gives training and decoding details.

\paragraph{Metrics and selection.}
We report official OMTG set tIoU, tF1@0.3/0.5, occurrence recall, count accuracy,
and EtF1, retaining all queries, including empty or invalid responses.
Official matching maximizes total IoU before thresholding; supplementary
micro diagnostics use thresholded maximum-cardinality matching.
Each model decodes once and selectors reuse its cached predictions.
For full-system OMTG results, one global threshold maximizes test tF1@0.5;
all metrics use that set. These are test-oracle operating points,
not deployment-validated thresholds; published baselines retain native rules.
The reported RL checkpoint was also selected after benchmark feedback.
Test labels enter offline metrics and selection feedback, never model inputs,
gradients, or rewards. Appendix~\ref{app:evaluation} specifies scoring and
threshold tie rules. Official OMTG scores average query-level metrics, so the
reported macro F1 is not computed by taking the harmonic mean of macro
precision and recall. Paired uncertainty checks resample source videos,
keeping queries from the same video together; these intervals condition on
the frozen model and candidate pool.

\subsection{Selection under equal return budgets}
\label{sec:confidence-score-comparison}

We rank the same 1,314 generation-order NMS survivors from the frozen RL model
on all 320 OMTG-Bench queries. \emph{Order} prioritizes earlier within-query
positions; \emph{Logp} uses mean interval-token log-probability;
\emph{Verifier} uses the frozen 2B SFT verifier's binary-normalized first-token
Yes probability; and \emph{Confidence} uses the span head from the original
decode. Only Verifier requires separate candidate-clip inference.
Each selector ranks across queries and receives the same global return count,
rounded down at each budget fraction; ties use generation index then query ID.
Generator outputs, intervals, and NMS are fixed, and GT only measures quality.
The quota is shared across the benchmark rather than imposed separately on
each query. A query may therefore receive several intervals or none, and
unanswered queries still contribute zero recall to the full denominator.

\begin{table}[!htbp]
  \centering
  \caption{\textbf{Confidence-score comparison at equal global return budgets.}
  Official OMTG-Bench Recall@0.5 (\%, $\uparrow$), averaged over all 320
  queries using the same 1,314 candidates. Budget denotes the retained
  fraction of candidates returned across queries.}
  \label{tab:global-budget-app}
  \small
  \setlength{\tabcolsep}{5pt}
  \renewcommand{\arraystretch}{1.10}
  \begin{tabular*}{\linewidth}{@{\extracolsep{\fill}}rcccc}
    \toprule
    Budget & Order & Logp & Verifier & Confidence \\
    \midrule
    10\% & 9.95 & 9.41 & 11.00 & \textbf{14.42} \\
    25\% & 26.48 & 22.37 & 26.31 & \textbf{31.12} \\
    50\% & 51.34 & 43.05 & 47.72 & \textbf{52.82} \\
    75\% & 64.76 & 61.00 & 62.45 & \textbf{65.30} \\
    \bottomrule
  \end{tabular*}
\end{table}

Confidence improves official query-macro recall over Order by 4.47 and 4.64
points at 10\% and 25\% retention, respectively
(Table~\ref{tab:global-budget-app}). The benefit is largest under tight budgets:
as more of the fixed pool is returned, selection has less room to help.
Paired 95\% source-video bootstrap intervals for these official macro-recall
gains are [2.30, 6.04] and [2.26, 7.09] points, respectively.

Appendix~\ref{app:fixed-pool} reports tie-sensitivity checks and supplementary
diagnostics of micro match counts and cross-query budget allocation.

\subsection{Threshold selection and transfer}
\label{sec:threshold-transfer}

\begin{figure}[t]
  \centering
  \includegraphics[width=\linewidth]{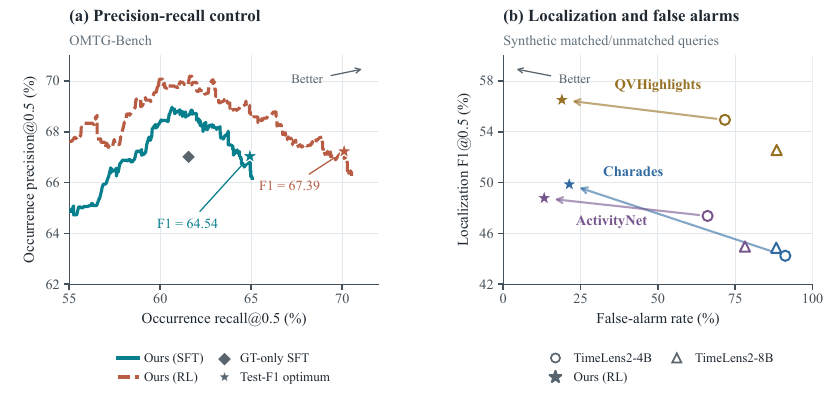}
  \caption{\textbf{Confidence-controlled selection and localization.}
  \textbf{(a)} OMTG-Bench precision--recall curves sweep a global threshold
  on each model's fixed candidates after generation-order NMS (0.3),
  without re-decoding. The high-precision/high-recall region is shown.
  The gray diamond is GT-only SFT at the official visual/512-token budget,
  without NMS or confidence selection.
  \textbf{(b)} Localization F1 versus false-alarm rate on the synthetic
  matched/unmatched splits of Table~\ref{tab:single-interval-unmatched}.
  Colors identify datasets; arrows connect native TimeLens2-4B to our RL
  model, with native TimeLens2-8B also shown. Higher F1 and lower false-alarm
  rate are better. In both panels, stars mark the global test-F1-optimal
  operating points reported in the corresponding tables.}
  \label{fig:sft-rl-pr}
\end{figure}

Figure~\ref{fig:sft-rl-pr}(a) sweeps global thresholds on each model's fixed
NMS pool. RL extends the high-recall end: retaining all survivors gives
70.51\% recall versus SFT's 65.06\%. Each curve reuses its own model's candidates;
separation between curves reflects both generator and scorer changes. For one fixed pool,
increasing $\tau$ produces nested accepted sets,
$S_{\tau_2}\subseteq S_{\tau_1}$ for $\tau_2\geq\tau_1$.
The return count therefore decreases or stays fixed; precision depends on
which correct and incorrect intervals the threshold removes.

Threshold transfer asks a different question. On 512 source-video-disjoint
calibration videos, we select thresholds meeting 80\% and 90\% empirical
micro precision, then apply them unchanged to OMTG-Bench. The achieved
precision falls to 71.71\% (654/912) and 86.89\% (391/450), respectively:
neither target transfers. The source cohort had been used in earlier project
analyses.
Appendix~\ref{app:threshold-transfer} reports calibration results and a
retrospective within-source variability check.

This experiment fits a threshold to an empirical precision criterion; it does
not simply interpret a confidence value of 0.8 as 80\% correctness. The two
transferred operating points retain micro recall of 55.75\% and 33.33\%,
respectively, so the higher precision comes with reduced coverage. The relative
contributions of threshold-selection variability, cohort composition, and
changes in the score--correctness relation remain unresolved.

Budgeted selection depends on score ordering, preserved by a common strictly
increasing transformation. Precision-target transfer additionally requires
predictions accepted at the source-selected threshold to maintain the required
correctness rate across cohorts.

\subsection{End-to-end grounding and component analysis}
\label{sec:rl-results}

\begin{table}[!htbp]
  \centering
  \caption{\textbf{Multi-occurrence grounding on OMTG-Bench.}
  Official metrics (\%). Native TimeLens2 uses the official visual/512-token
  budget; published baselines retain source protocols
  \citep{zhu2026timelens2,xu2026omtg}.
  $^\dagger$: one global test-tF1@0.5-optimal threshold.}
  \label{tab:main-results}
  \small
  \setlength{\tabcolsep}{3.7pt}
  \renewcommand{\arraystretch}{1.08}
  \begin{tabular*}{\textwidth}{@{\extracolsep{\fill}}lccccc@{}}
    \toprule
    Model & C-Acc & tF1@0.3 & tF1@0.5 & tIoU & EtF1 \\
    \midrule
    Seed-1.8 & 38.12 & 67.13 & 54.67 & 56.81 & 28.04 \\
    Gemini 2.5 Pro & 50.94 & 55.72 & 43.57 & 43.24 & 27.80 \\
    \midrule
    Qwen3-VL-4B & 0.31 & 37.07 & 26.75 & 30.42 & 0.21 \\
    TimeLens-8B & 0.00 & 39.14 & 32.76 & 32.38 & 0.00 \\
    TimeLens2-4B & 18.75 & 53.02 & 43.89 & 48.73 & 14.29 \\
    TimeLens2-8B & 16.25 & 47.14 & 38.19 & 47.08 & 12.86 \\
    OMTG-4B &
      \textbf{55.63} & 73.46 & 65.40 & 61.24 & \textbf{43.65} \\
    \midrule
    Ours (SFT)$^\dagger$ & 51.88 & 72.40 & 64.54 & 60.75 & 40.26 \\
    Ours (RL)$^\dagger$ &
      53.75 & \textbf{75.92} & \textbf{67.39} & \textbf{63.24} & 42.64 \\
    \bottomrule
  \end{tabular*}
\end{table}

Our SFT and RL models share the backbone, visual budget, parser, NMS, and
one-decode inference, each with its matching head. RL gains 3.52 points in
tF1@0.3, 2.85 in tF1@0.5, and 2.49 in tIoU over SFT
(Table~\ref{tab:main-results}).
OMTG-4B retains stronger count accuracy and EtF1.

\begin{table}[!htbp]
  \centering
  \caption{\textbf{Candidate coverage and successive selection stages.}
  Official Recall@0.5 and tF1@0.5 (\%) on fixed decoded pools.
  NMS uses generation order at tIoU 0.3; $^\dagger$ denotes each model's
  global test-F1-optimal confidence threshold.}
  \label{tab:confidence-ablations}
  \small
  \setlength{\tabcolsep}{5pt}
  \begin{tabular*}{\linewidth}{@{\extracolsep{\fill}}lrrrrrr@{}}
    \toprule
    & \multicolumn{2}{c}{Raw} & \multicolumn{2}{c}{NMS} &
      \multicolumn{2}{c}{NMS + confidence$^\dagger$} \\
    \cmidrule(lr){2-3}\cmidrule(lr){4-5}\cmidrule(l){6-7}
    Model & Recall & tF1 & Recall & tF1 & Recall & tF1 \\
    \midrule
    SFT & 67.85 & 34.65 & 65.06 & 64.17 & 64.92 & 64.54 \\
    RL & 73.64 & 41.39 & 70.51 & 67.07 & 70.10 & 67.39 \\
    \bottomrule
  \end{tabular*}
\end{table}

Table~\ref{tab:confidence-ablations} shows raw recall rising from 67.85 to 73.64
as RL reduces the mean candidate count from 11.18 to 8.99 (about 20\%);
the coverage gain thus does not rely on more candidates on average.
NMS accounts for most of the raw-to-final F1 gain; confidence filtering on
the fixed RL pool adds 0.32 points (paired 95\% interval [$-$0.14, 0.76]), alongside
the budgeted-selection gains in Table~\ref{tab:global-budget-app}.
Confidence-sorted NMS did not improve performance. The GT-only system
reference matches the inference budget but differs in initialization,
data, and prompt.

On fixed L200 candidates and decoder states, the online head improves post-NMS
AUROC from 0.754 to 0.818 over the original SFT head
(Appendix~\ref{app:head-crossover}). It raises 10\%-budget recall by 1.62 points
(95\% interval [0.14, 2.83]); gains at 25\% and in test-oracle F1 remain
uncertain. Better candidate-correctness discrimination can thus support
tight-budget selection without a clear final-set F1 gain, consistent with the
gap between local overlap labels and marginal set utility
(Section~\ref{sec:confidence-learning}).

\subsection{Ranking and rejection on single-interval datasets}
\label{sec:single-interval-unmatched}

Each matched query in the Charades, ActivityNet, and QVHighlights TimeLens
splits is paired with one cross-video mismatch. These synthetic negatives are
not official annotations or exhaustively human-verified absences. All prompts
permit \texttt{[]}. Our frozen models retain the highest-confidence NMS
survivor and reject it below one benchmark-wide threshold; TimeLens2 baselines
retain their native output rules.

\begin{table}[!htbp]
  \centering
  \caption{\textbf{Localization and rejection on balanced matched/synthetic-mismatched splits.}
  All prompts allow empty outputs. $^\dagger$ denotes one benchmark-wide
  test-F1-optimal threshold; native baselines retain their original output
  rules. F1@0.5 and false-alarm rate (FAR) are percentages.}
  \label{tab:single-interval-unmatched}
  \small
  \setlength{\tabcolsep}{4pt}
  \renewcommand{\arraystretch}{1.10}
  \begin{tabular*}{\textwidth}{@{\extracolsep{\fill}}lcccccc@{}}
    \toprule
    & \multicolumn{2}{c}{Charades} & \multicolumn{2}{c}{ActivityNet} & \multicolumn{2}{c}{QVHighlights} \\
    \cmidrule(lr){2-3}\cmidrule(lr){4-5}\cmidrule(l){6-7}
    Method & F1@0.5 $\uparrow$ & FAR $\downarrow$ & F1@0.5 $\uparrow$ & FAR $\downarrow$ & F1@0.5 $\uparrow$ & FAR $\downarrow$ \\
    \midrule
    TimeLens2-4B, native & 44.24 & 91.17 & 47.38 & 66.04 & 54.95 & 71.64 \\
    TimeLens2-8B, native & 44.87 & 88.25 & 44.97 & 78.13 & 52.57 & 88.38 \\
    Ours (SFT)$^\dagger$ & 46.81 & 36.01 & 48.42 & 15.22 & \textbf{57.67} & \textbf{16.87} \\
    Ours (RL)$^\dagger$ & \textbf{49.87} & \textbf{21.35} & \textbf{48.78} & \textbf{13.24} & 56.52 & 18.95 \\
    \bottomrule
  \end{tabular*}
\end{table}

In this full-system rejection comparison, native TimeLens2-4B retains
false-alarm rates of 66.04--91.17\% on these mismatches. At test-F1-optimal
thresholds, our RL system reduces this range to 13.24--21.35\% and improves
F1 by 1.40--5.63 points (Table~\ref{tab:single-interval-unmatched}).
ActivityNet and QVHighlights gains trade recall for precision. SFT already
reduces false alarms and exceeds RL on QVHighlights, so the additional benefit
of RL varies by dataset.

A separate fixed-pool analysis on the matched subsets compares returning the
first raw interval with returning the highest-confidence interval, without
NMS or thresholding. On Charades, confidence improves mIoU by 2.02 points
with a paired 95\% interval of [1.39, 2.68]. ActivityNet and QVHighlights
have positive point estimates but intervals that include zero
(Appendix~\ref{app:single-interval}). This supports within-query ranking on
Charades while leaving its consistency across datasets unresolved.

\FloatBarrier
\section{Related Work}
\label{sec:related-work}

\paragraph{Video temporal grounding.}
Discriminative methods learn matching or saliency scores
\citep{gao2017tall,lei2021qvhighlights,lin2023univtg,moon2023qddetr}; generative approaches
produce timestamps through language decoding or dedicated interval prediction
\citep{huang2024vtimellm,ren2024timechat,wang2024groundedvideollm,wu2025numpro,pramanick2025edvtg}.
Time-R1 and TimeLens study RL-based grounding and training recipes
\citep{wang2025timer1,zhang2026timelens}, while OMTG and TimeLens2 develop
set-valued generation \citep{xu2026omtg,zhu2026timelens2}.
TimeExpert also generates saliency tokens for highlight detection
\citep{yang2025timeexpert}. Our score targets each decoded interval's
localization correctness for acceptance decisions. MetaVideoAgent evolves video-agent pipelines, including evidence localization, for long-form video understanding \citep{cui2026metavideoagent}.

\paragraph{Grounding reliability and rejection.}
RaTSG, OpenVMR, and Moment of Untruth address absent-query rejection
\citep{dong2024ratsg,fang2024openvmr,flanagan2025untruth}; Generalized Video Moment Retrieval
supports both multiple and no-target queries \citep{qin2025gvmr}.
RA-RFT trains generative grounders to refuse semantically similar but
irrelevant queries through GRPO \citep{lee2026rarft}; HRVTG adapts grounders
at test time using counterfactual probes \citep{yi2026hrvtg}.
Event verification tuning addresses localization--verification inconsistencies
\citep{jung2025consistency}; CAVE aligns boundary evidence through RL and
competence-aware gating \citep{jia2026cave}.
Our method supports rejection by selecting intervals from a completed
grounding response with a frozen confidence head.

\paragraph{Confidence and selective prediction.}
ConfidNet learns auxiliary confidence from model features
\citep{corbiere2019confidnet}, and hidden-state classifiers predict statement
truthfulness in LLMs \citep{azaria2023internal}.
For temporal grounding, URPA uses rollout variance to weight adaptation
rewards \citep{hu2025urpa}.
Selective prediction controls acceptance and risk \citep{geifman2019selectivenet},
whereas calibration concerns agreement between scores and empirical correctness
\citep{guo2017calibration}. Language-model approaches include verbalized
confidence \citep{tian2023calibration}, uncertainty across sampled answers
\citep{kuhn2023semanticuncertainty}, and hidden-state entropy probes
\citep{kossen2024semanticentropyprobes}. Building on learned confidence
readouts, we train interval scores with offline visual-support targets and
online temporal-overlap labels for selection from one grounding response.

\section{Limitations}
\label{sec:discussion}

This work explores a confidence-aware paradigm for TVG, in which confidence
is explicitly learned during training and used to guide inference. Our current
confidence head represents one practical realization of this direction: a
lightweight MLP over pooled decoder states, adapted online with a standard
binary cross-entropy objective. Although this design demonstrates the utility
of incorporating confidence into TVG, it explores only a limited part of the
architectural and optimization space. In particular, confidence supervision
updates the readout without directly shaping the underlying representations.
Confidence-based selection also depends on the available candidate pool and
cannot recover missing or NMS-suppressed proposals. Future work will investigate
more native integration of confidence into temporal representation learning
and generation, through architectures and training objectives that jointly
learn where to ground and how certain the model should be. Establishing
probability calibration and its robustness across domains remains a further
research direction. Localization confidence does not imply robustness to meta-adaptive multimodal jailbreaks \citep{cui2026fully}, which we do not evaluate.

\section{Conclusion}
\label{sec:conclusion}

We introduced interval-level confidence as a decision interface for generative
video temporal grounding. By separating candidate generation from acceptance,
our approach allows a single decoded response to support ranking,
threshold-based selection, and rejection without an inference-time verifier.
Fixed-pool experiments on OMTG-Bench show improved recall under equal return
budgets, with the largest gains at tight budgets. These results highlight that
generation and selection are distinct capabilities: the value of a candidate
pool also depends on how its intervals are prioritized and accepted. Explicit
confidence makes the output policy adjustable after decoding, allowing the
same candidates to serve different return budgets and acceptance requirements.

\clearpage
\subsubsection*{AI use statement}
Generative AI tools were used primarily to polish the manuscript's language,
develop editable figure code, and refine LaTeX formatting.

\bibliographystyle{iclr2027_conference}
\bibliography{references}

\clearpage
\appendix
\raggedbottom
\section{Supplementary Selection Results and Statistical Checks}
\label{app:fixed-pool}
\label{app:single-interval}

\paragraph{Official recall at a global acceptance budget.}
Section~\ref{sec:confidence-score-comparison} and
Table~\ref{tab:global-budget-app} compare four selectors at equal global return
budgets. Selected intervals are restored to generation order before the official
parser. Official Hungarian matching maximizes total IoU, then counts pairs
with tIoU $\geq0.5$; recall averages each query's matched fraction over all
320 queries. This differs from the RL reward's thresholded maximum-cardinality
matching, so all-survivor recall is not a formal upper bound. The point
estimates condition on the benchmark-selected checkpoint and candidate pool.

For Table~\ref{tab:global-budget-ci}, we reuse the 3,000 paired source-video
bootstrap draws of the micro analysis below. Each draw samples 287 videos with
replacement, retaining all queries from each sampled video. Each scorer
re-ranks the replicated NMS candidates and returns
$\lfloor fN_b\rfloor$ intervals, where $f$ is the budget fraction and $N_b$
is the resampled pool size. We apply the same official parser and evaluator
to each query copy and average recall over all query copies, including
unanswered ones. The paired percentile intervals therefore use the same
metric and budget policy as Table~\ref{tab:global-budget-app}.

\begin{table}[H]
  \centering
  \caption{\textbf{Official recall differences at tight global budgets.}
  Confidence minus each baseline in query-macro Recall@0.5 (percentage
  points), with paired 95\% percentile intervals.}
  \label{tab:global-budget-ci}
  \small
  \setlength{\tabcolsep}{4pt}
  \begin{tabular*}{\linewidth}{@{\extracolsep{\fill}}rccc@{}}
    \toprule
    Budget & $\Delta$ vs.\ Order & $\Delta$ vs.\ Logp & $\Delta$ vs.\ Verifier \\
    \midrule
    10\% & +4.47 [2.30, 6.04] & +5.01 [3.26, 6.36] & +3.42 [1.36, 4.56] \\
    25\% & +4.64 [2.26, 7.09] & +8.75 [5.86, 11.29] & +4.81 [2.33, 7.16] \\
    \bottomrule
  \end{tabular*}
\end{table}

\paragraph{Micro match counts and allocation diagnostics.}
At 50\% retention (657 intervals), Confidence has 528 matched intervals and
80.37\% micro precision, versus 454/69.10\% for Order and 497/75.65\% for
Verifier. It answers 257 of 320 queries, versus Verifier's 275: more total
matches are concentrated on fewer queries. These maximum-cardinality micro
counts are a diagnostic distinct from official query-macro recall.

At 10\%, 1,000 random cross-query Order tie breaks give 84--99 matches in the
central 95\% (mean 91.58), versus 87 with fixed ID ties and 124 for Confidence;
this is tie sensitivity, not a confidence interval. In 3,000 paired
source-video bootstrap draws, each scorer re-ranks and receives the same
recomputed quota. Exploratory 95\% intervals for Confidence-minus-Order and
Confidence-minus-Verifier \emph{micro-precision} differences are [18.80, 36.51]
and [$-$0.85, 15.27] points at 10\%; [12.50, 22.94] and [0.62, 10.77]
at 25\%; and [8.87, 15.31] and [1.64, 8.12] at 50\%.
The 10\% Verifier comparison remains inconclusive for micro precision even
though its query-macro recall interval is positive. Both bootstrap analyses
are exploratory, exclude checkpoint-selection and training-seed uncertainty,
and have no multiple-comparison correction.

\paragraph{Ranking is useful beyond generation order.}
A model may emit multiple plausible boundaries even when evaluation requires
one answer. We compare taking the first valid interval with taking the
highest-confidence interval from the \emph{same} response. Ties use generation
order; neither rule uses NMS or a confidence threshold. The generator and its
matching head remain frozen throughout all three datasets.

\begin{table}[H]
  \centering
  \caption{\textbf{Single-answer ranking on fixed decoded pools.}
  Complete matched subsets of the mixed-query TimeLens runs; mIoU and R@1 at tIoU 0.5 are
  percentages. The paired mIoU changes and 95\% intervals use 2,000
  source-video bootstrap samples. Both selectors reuse the saved responses
  to the prompt allowing \texttt{[]}; invalid or empty answers have zero IoU.}
  \label{tab:single-interval-app}
  \small
  \setlength{\tabcolsep}{4pt}
  \begin{tabular*}{\linewidth}{@{\extracolsep{\fill}}lrcccc@{}}
    \toprule
    & & \multicolumn{2}{c}{mIoU} & & R@1, 0.5 \\
    Dataset & Queries & First & Confidence & $\Delta$ mIoU [95\% CI] & First $\to$ Conf. \\
    \midrule
    Charades & 3,363 & 47.89 & 49.91 & +2.02 [1.39, 2.68] & 55.16 $\to$ 57.21 \\
    ActivityNet & 4,500 & 45.96 & 46.40 & +0.44 [$-$0.21, 1.11] & 50.76 $\to$ 51.33 \\
    QVHighlights & 1,541 & 57.26 & 57.95 & +0.68 [$-$0.38, 1.69] & 61.19 $\to$ 62.10 \\
    \bottomrule
  \end{tabular*}
\end{table}

Confidence improves the point estimates in Table~\ref{tab:single-interval-app},
but only the Charades mIoU interval excludes zero. Generation order leaves
useful selection information unused on that dataset; the smaller gains on
ActivityNet and QVHighlights remain uncertain.

\section{Threshold Transfer and Component Analysis}
\label{app:threshold-transfer}

\paragraph{Threshold transfer.}
We also fit global thresholds on 512 source-video-disjoint calibration videos,
selecting the largest NMS-survivor set that meets 80\% or 90\% empirical
micro precision, with minimum acceptance and video-coverage requirements.
The calibration set is disjoint from project SFT, RL, and OMTG videos, but was
used in earlier project analyses; it is not a fresh confirmation set.
Calibration precision was 80.03\% (1,190/1,487) and 90.09\% (845/938).
Applying those frozen thresholds once to OMTG gives 71.71\% (654/912) and
86.89\% (391/450) micro precision, with micro recall 55.75\% and 33.33\%,
respectively. Neither empirical target transfers.
In a retrospective five-fold video-held-out check within the same source
cohort, pooled held-out precision is 80.00\% (1,196/1,495) and 89.94\%
(840/934); only 3/5 and 2/5 folds meet their respective targets.

\subsection{NMS and confidence selection}
\label{app:nms-order}

\begin{figure}[H]
  \centering
  \includegraphics[width=\linewidth]{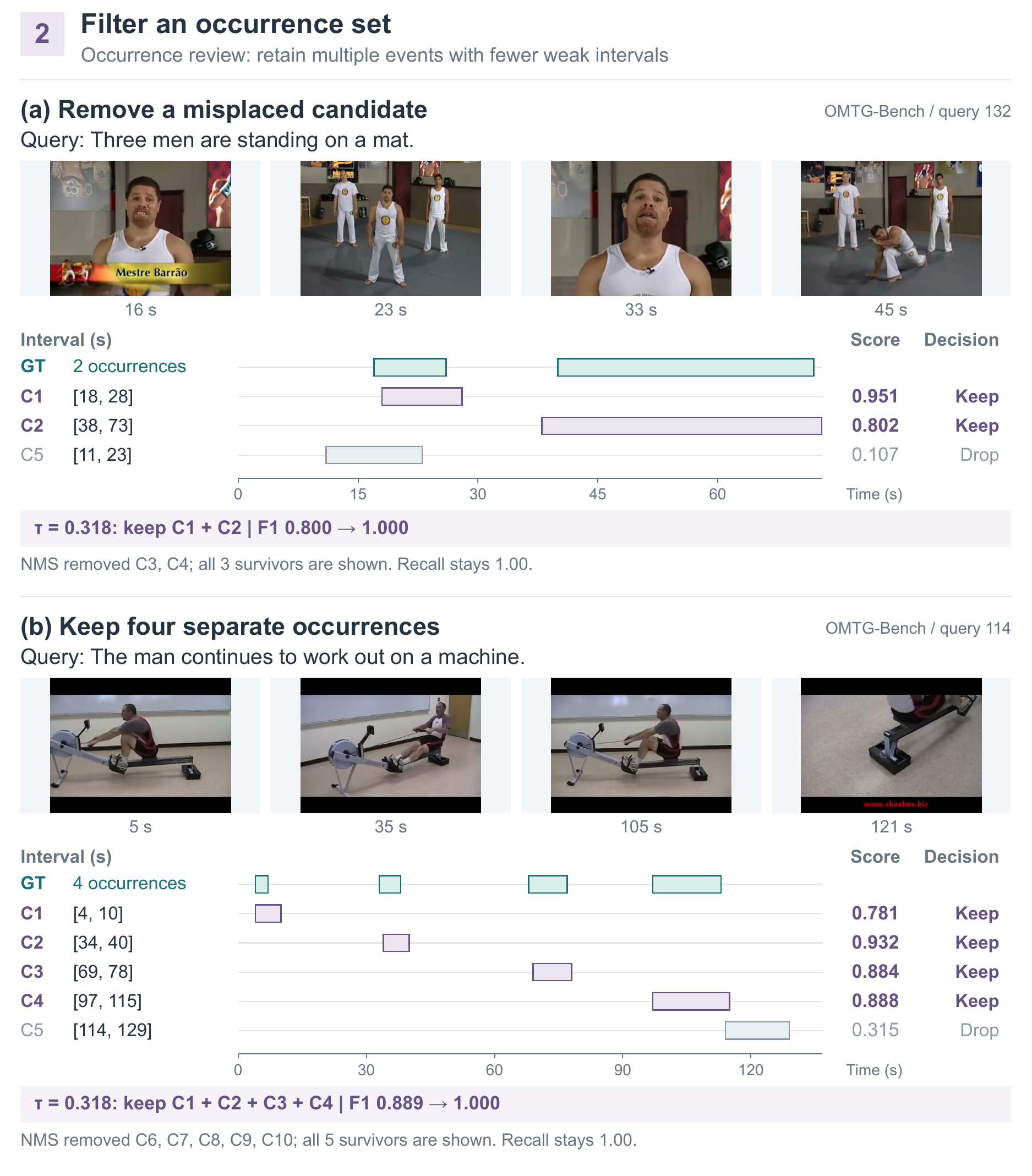}
  \caption{\textbf{Set filtering: review repeated events with fewer low-quality intervals.}
  Both OMTG-Bench cases use generation-order NMS at tIoU 0.3 followed by
  that variant's global test-oracle threshold, $\tau\approx0.318$.
  Every NMS survivor is shown, with its original candidate index.
  \textbf{(a)} A misplaced interval survives NMS but is removed by confidence,
  retaining both annotated occurrences. \textbf{(b)} Four workout intervals
  are retained while the extra $[114,129]$ fragment is removed.
  F1 rises from 0.800 and 0.889 to 1.000, respectively; recall remains
  1.000 in both cases. The same threshold supports different answer counts.}
  \label{fig:confidence-case-sets}
\end{figure}

\paragraph{Redundancy removal and acceptance control are complementary.}
NMS uses overlap and generation order; confidence uses an interval's decoder
features. Table~\ref{tab:nms-confidence-app} applies the two rules in sequence,
holding each model's decoded pool fixed. On RL candidates, confidence raises
precision from 66.25 to 67.23 and F1 from 67.07 to 67.39, while recall changes
from 70.51 to 70.10. The paired F1 change is +0.32 points, with a 95\%
source-video bootstrap interval of [$-$0.14, 0.76] (3,000 draws;
threshold held fixed), so the incremental F1 benefit is not established.
The historical cases in Figure~\ref{fig:confidence-case-sets} illustrate
how the two selectors can make different decisions.

\begin{table}[H]
  \centering
  \caption{\textbf{Confidence selection after geometric suppression.}
  Official OMTG-Bench metrics (\%), averaged over all 320 queries.
  Confidence uses each model's global F1-optimal test-oracle threshold;
  NMS alone retains every survivor.}
  \label{tab:nms-confidence-app}
  \small
  \setlength{\tabcolsep}{5pt}
  \begin{tabular*}{\linewidth}{@{\extracolsep{\fill}}llccc@{}}
    \toprule
    Model & Selection & Precision@0.5 & Recall@0.5 & tF1@0.5 \\
    \midrule
    SFT & NMS & 66.17 & 65.06 & 64.17 \\
    SFT & NMS + confidence & 67.04 & 64.92 & 64.54 \\
    \midrule
    RL & NMS & 66.25 & 70.51 & 67.07 \\
    RL & NMS + confidence & 67.23 & 70.10 & 67.39 \\
    \bottomrule
  \end{tabular*}
\end{table}

\paragraph{NMS order control.}
On cached final-model candidates, confidence-sorted NMS lowers tF1@0.5 from
67.39 to 66.61 at the same test-oracle threshold. Exploratory five-fold
source-video-grouped threshold selection gives 67.37 versus 66.41
(paired bootstrap difference [$-$1.76, $-$0.21] points). Both comparisons
reuse the inspected OMTG cohort. We retain generation-order NMS based on
these aggregate results.

\subsection{Full system and candidate-generation reference}
\label{app:components}

\begin{table}[H]
  \centering
  \caption{\textbf{Candidate generation and final selection on OMTG-Bench.}
  All rows use the official visual budget and greedy decoding with 512 output
  tokens on all 320 queries (1,173 GT intervals); metrics are percentages.
  The upper block shows raw outputs; the lower applies generation-order
  NMS (0.3) and confidence selection.
  $^\dagger$ denotes one global F1-optimal test-oracle threshold, shared by
  all metrics in that row.}
  \label{tab:full-components-app}
  \small
  \setlength{\tabcolsep}{4pt}
  \renewcommand{\arraystretch}{1.12}
  \begin{tabular*}{\linewidth}{@{\extracolsep{\fill}}lccccc@{}}
    \toprule
    Variant & \shortstack{Recall\\@0.5} & C-Acc &
      \shortstack{tF1\\@0.5} & tIoU & EtF1 \\
    \midrule
    \multicolumn{6}{@{}l}{\textit{Candidate generation: raw outputs}} \\
    GT-only SFT & 61.55 & 48.13 & 62.59 & 58.67 & 38.08 \\
    Ours (SFT), raw & 67.85 & 0.00 & 34.65 & 53.35 & 0.00 \\
    Ours (RL), raw & \textbf{73.64} & 0.31 & 41.39 & 57.21 & 0.01 \\
    \midrule
    \multicolumn{6}{@{}l}{\textit{Final selection: NMS + confidence selection}} \\
    Ours (SFT)$^\dagger$ &
      64.92 & 51.88 & 64.54 & 60.75 & 40.26 \\
    Ours (RL)$^\dagger$ &
      70.10 & \textbf{53.75} & \textbf{67.39} & \textbf{63.24} & \textbf{42.64} \\
    \bottomrule
  \end{tabular*}
\end{table}

The GT-only reference returns 3.65 candidates per query and reaches 61.55
Recall@0.5. Our raw SFT and RL pools return 11.18 and 8.99 candidates, with
67.85 and 73.64 recall. After NMS and confidence selection, their mean returned
counts are 3.89 and 3.74. The GT-only run matches the official visual and
512-token inference budget but differs in training initialization, data
composition, and prompt; these rows do not isolate candidate-target construction.

\subsection{Head adaptation on fixed candidates and decoder states}
\label{app:head-crossover}

We rescore the final step-200 actor's 2,878 cached candidates and identical
span features with the original SFT head and its matching online RL head.
Both heads remain frozen, with no new generator decoding or optimizer steps.
Generation-order NMS at 0.3 leaves the same 1,314 candidates from 320 queries
and 287 source videos. Candidate AUROC uses the local label
$\mathbf{1}[\max_j\operatorname{tIoU}(I_i,Y_j)\geq0.5]$ after NMS.

\begin{table}[H]
  \centering
  \caption{\textbf{Head crossover on the same L200 candidates and features.}
  Recall is official query-macro Recall@0.5 at the indicated global return
  budget. Each head's tF1 uses its own global test-F1-optimal threshold;
  these are diagnostic operating points. Recall and tF1 are percentages.}
  \label{tab:head-crossover}
  \small
  \setlength{\tabcolsep}{4pt}
  \begin{tabular*}{\linewidth}{@{\extracolsep{\fill}}lcccc@{}}
    \toprule
    Head & NMS AUROC & Oracle tF1@0.5 & Recall, 10\% & Recall, 25\% \\
    \midrule
    Original SFT & 0.754 & 67.18 & 12.80 & 29.74 \\
    Online RL & 0.818 & 67.39 & 14.42 & 31.12 \\
    \bottomrule
  \end{tabular*}
\end{table}

At 10\% and 25\% budgets (131 and 328 returns), the online-minus-SFT recall
differences are +1.62 [0.14, 2.83] and +1.38 [$-$0.59, 3.11] points,
respectively. These paired 95\% percentile intervals reuse the 3,000
source-video draws from Appendix~\ref{app:fixed-pool}, reallocate the global
budget in each draw, and apply the same official evaluator to every query
copy, including unanswered ones. The 10\% gain is supported on this frozen
pool; the 25\% difference remains uncertain.

For test-oracle F1, the thresholds are 0.257 for the SFT head and 0.184 for
the online head. The difference is +0.204 points, with a paired 95\%
source-video bootstrap interval of [$-$0.063, 0.478] (5,000 draws,
thresholds held fixed). Across queries, the online head wins on 16, loses
on 6, and ties on 298. Stronger candidate discrimination and a tight-budget
recall gain therefore coexist with an unresolved final-set F1 gain.
This is consistent with the distinction between individual overlap labels
and marginal set utility in Section~\ref{sec:confidence-learning}.
The actor itself was trained with online adaptation; this inference control
does not isolate head-training choices or establish online adaptation as
necessary for final F1. All intervals condition on the benchmark-selected
checkpoint and fixed pool, excluding checkpoint-selection and training-seed
uncertainty; no multiple-comparison adjustment is applied.

\section{Reproducing the Reported Models}
\label{app:implementation}

\paragraph{Model identity and training order.}
The SFT row uses the completed 1,888-step model and its jointly trained
confidence head. The RL row uses a two-stage continuation of that merged
generator. Steps 1--100 use $0.5\operatorname{Recall}_{0.5}(P,\mathcal{Y})
+0.5\operatorname{tIoU}(S_{\tau_{\mathrm{train}}},\mathcal{Y})$ with fixed
$\tau_{\mathrm{train}}\approx0.183$. This value was selected for the final SFT
model by a global threshold sweep on 512 calibration examples, maximizing
query-macro occurrence F1@0.5 after generation-order NMS at 0.3 under the
earlier 120-frame calibration protocol. It remains fixed throughout
steps 1--100. Steps 101--200 use
Eq.~\eqref{eq:set-localization-reward}, adding selected-set F1 and choosing
one threshold across the 512 training responses. Ties favor higher recall,
then a lower threshold. At step 101, actor LR
changes from $10^{-6}$ to $2\times10^{-7}$; optimizer moments, scheduler,
RNG, and data position are preserved. All quantitative RL rows use this
step-200 actor and its matching head after 150 online updates. The SFT head
stays frozen for steps 1--50; online updates begin at step 51. Both stages
use only OMTG RL data, with no additional training-data mixture.

\begin{table}[H]
  \centering
  \caption{\textbf{Training settings for the main-table models.}}
  \label{tab:training-settings-app}
  \small
  \setlength{\tabcolsep}{5pt}
  \renewcommand{\arraystretch}{1.08}
  \begin{tabular*}{\linewidth}{@{\extracolsep{\fill}}lp{0.64\linewidth}@{}}
    \toprule
    Component & Setting \\
    \midrule
    Base model & TimeLens2-4B; hidden size 2,560 \\
    Confidence head & LayerNorm, Linear(2560, 256), GELU, dropout 0.1, Linear(256, 1), sigmoid \\
    SFT adaptation & LoRA rank 32, alpha 64, dropout 0.05; fresh head \\
    SFT optimizer & AdamW; generator LR $2\times10^{-5}$, head LR $2\times10^{-4}$; cosine decay, 3\% warmup \\
    SFT schedule & 60,405 units; global batch 64; two epochs, 1,888 steps \\
    RL actor & Full-parameter updates; 200 outer steps; LR $10^{-6}$ for steps 1--100, then $2\times10^{-7}$ \\
    RL sampling & 64 queries $\times$ 8 responses; temperature 1.0, top-$p$ 0.8, top-$k$ 20 \\
    RL policy loss & Query minibatch 16; mean-centered advantages without standard-deviation normalization; sequence-mean token-mean aggregation \\
    Policy clipping & Ratio interval $[0.8,1.285]$; dual-clip coefficient 10 \\
    Regularization & No KL reward, KL loss, or entropy bonus \\
    Training selector & Generation-order NMS 0.3; fixed threshold for steps 1--100, then one batch-shared F1-optimal threshold \\
    Online head & FP32 AdamW; LR $10^{-4}$; weight decay 0.01; gradient norm cap 1 \\
    \bottomrule
  \end{tabular*}
\end{table}

\paragraph{Two supervised views.}
SFT uses 60,405 OMTG/TimeLens2 training units representing 60,052 distinct
video--query pairs and 49,766 source videos. Each epoch contains 219,000 GT
anchors, 73,284 real alternatives, and 328,188 synthetic boundary alternatives.
GT anchors are retained before alternatives, with at most two alternatives
per GT, 32 total candidates, and 480 completion tokens. Real alternatives
require maximum GT tIoU at least 0.5; synthetic ones require 0.6. Added
alternatives cannot duplicate or overlap an existing interval at tIoU 0.85
or above. Alternative slots are allocated across occurrences in rounds.
Real proposals are used first, and synthetic perturbations fill remaining slots.
The target closes only after all retained anchors and alternatives, without
fixed-count padding or low-IoU filler.
Queries with inconsistent durations or too many GT intervals are excluded
before target construction rather than partially dropping their annotations.

The confidence view replays one fixed candidate sequence per training unit,
chosen by the first source identifier in sorted order. It contains 594,620
candidate occurrences per epoch: 152,824 matched proposals, 66,340 unmatched
proposals, 66,176 GT completions, and 309,280 duration-based negatives.
The corresponding 582,900 unique verifier scores remain continuous.
Both epochs reuse the same targets and replays. Replay has zero LM loss,
and its pooled interval features are detached. Synthetic generator alternatives
do not inherit fabricated verifier scores. The head starts from random
initialization with seed 20260902; no separate frozen-generator head-fitting
stage is used for this SFT model.

The offline verifier is a separately fine-tuned Qwen3.5-2B model
(checkpoint \texttt{candidate-verifier-scaleup-v2-final-step-1159}), trained
on 74,155 candidate clips, with 2,800 independent calibration clips.
It answers a Yes/No
question about the query and candidate clip; the soft target is the
binary-normalized first-answer-token probability of Yes.
The verifier supplies offline SFT targets and the comparison baseline in
Table~\ref{tab:global-budget-app}; it is not used for RL label construction
or inference by our grounding model.

For offline target generation and the Table~\ref{tab:global-budget-app}
baseline, the user message contains a candidate-local video clip followed by
the text below. Times are clip-relative seconds, formatted to three decimal
places.
\begin{quote}
\small\ttfamily\raggedright
Judge one temporal-grounding candidate using the video evidence.\\
Target event: \{query\}\\
Clip duration: \{duration:.3f\} seconds\\
Candidate interval within this clip: \{start:.3f\} - \{end:.3f\} seconds

Does this interval correctly localize an occurrence of the target event?
Answer exactly Yes or No. Do not explain.
\end{quote}
The verifier's own SFT uses the same text without the
\texttt{Clip duration} line.

\paragraph{SFT loss details.}
Let $\ell_{\mathrm{GT}}$, $\ell_{\mathrm{alt}}$, and $\ell_{\mathrm{fmt}}$
be token-averaged cross-entropy losses within their respective groups.
The generator objective is
\begin{equation}
  \mathcal{L}_{\mathrm{gen}}=
  \begin{cases}
    0.6\ell_{\mathrm{GT}}+0.3\ell_{\mathrm{alt}}+0.1\ell_{\mathrm{fmt}},
      & \text{with alternatives},\\
    0.9\ell_{\mathrm{GT}}+0.1\ell_{\mathrm{fmt}},
      & \text{otherwise}.
  \end{cases}
  \label{eq:generator-sft}
\end{equation}
For confidence replay, $\mathcal{L}_{\mathrm{pair}}$ averages
$\log(1+\exp[-(a_i-a_j)])$ over within-query pairs with $z_i-z_j\geq0.1$.
It is zero if no eligible pair exists. Its weight within the head loss is
0.25, and the head loss weight in the joint SFT objective is also 0.25
(Eqs.~\eqref{eq:head-objective}--\eqref{eq:supervised-objective}).

\paragraph{Online updates.}
RL uses 8,424 OMTG training queries and a source-video-disjoint 128-query
internal diagnostic split. At each step, the current head scores the sampled
intervals and the reward remains fixed through the actor update. From step 51,
the head then takes one BCE update on valid NMS survivors, labeled by maximum
training-GT tIoU at least 0.5. Losses are averaged over candidates within a
rollout, then over valid rollouts across eight ranks. Detached features are
reused from the actor's old-log-probability forward pass. Only head parameters
receive the confidence gradient; the updated head scores the next batch.
The head is in evaluation mode for scoring and uses dropout only during its
own supervised update. Generator and head are saved and evaluated together.

\paragraph{Visual and decoding budgets.}
SFT uses 2 fps, at most 120 frames, and at most 64 visual tokens per frame.
RL and OMTG evaluation use 2 fps, \texttt{min\_pixels}=2048,
\texttt{total\_pixels}=8,388,608, and a 16,384-token context.
TimeLens single-interval evaluation uses 4 fps, at most 2,048 frames,
784--200,704 pixels per frame, a 50,176,000-pixel video budget, and a
131,072-token context. Greedy evaluation permits 512 new tokens and
32 unique candidates, with the same trained candidate-generation prompt.
Repetition guards close the response after three consecutive duplicate
candidates or 40 total candidates. Confidence is read from the states of
consumed interval tokens in that same decode; there is no second video
encoding, teacher-forced inference replay, or generated confidence token.
The strict parser preserves original candidate indices for score alignment;
invalid boundaries are never clipped, swapped, or invented.

\section{Evaluation Protocols}
\label{app:evaluation}
\label{app:unmatched-protocol}

\paragraph{Shared evaluation design.}
Each model decodes once, and selectors reuse its cached predictions.
Generation-order NMS precedes threshold-based selection, whereas the
single-answer ranking analysis uses raw candidates.
OMTG uses official metrics with all queries in the
denominator; malformed outputs are empty and invalid boundaries are filtered
without changing timestamps. Reported global F1-optimal thresholds use test
labels and are test-oracle operating points. Native baselines retain their
original output rules. For the main OMTG threshold scan, ties in official
tF1@0.5 use tIoU, EtF1, fewer retained candidates, then a higher threshold.
All metrics in each reported row use the same selected set. Both main models
have zero invalid or truncated OMTG responses.

\paragraph{Native TimeLens2 baselines.}
We rerun the released TimeLens2-4B and 8B checkpoints with their original README prompt
on all 320 OMTG queries, using the official visual budget, a 16,384-token
context, and greedy decoding with 512 output tokens. The common repetition
and candidate-count guards apply; raw outputs are evaluated without NMS or
confidence selection against all 1,173 official GT intervals. Invalid or
incomplete responses remain in the denominator as empty predictions.
The 4B and 8B reruns use FlashAttention 2 and SDPA, respectively.

\paragraph{Fixed-pool comparisons.}
Single-answer ranking compares first versus highest-confidence intervals from
the same saved candidates. Paired bootstrap resamples source videos, preserving
queries from each video together.

\paragraph{Rejection on synthetic unmatched queries.}
Each matched query is paired with a cross-video mismatch; prompts permit
\texttt{[]}. We retain the highest-confidence NMS survivor and reject it below
the global threshold. At tIoU 0.5, an incorrect positive counts as both FP
and FN; a nonempty or invalid unmatched response counts as a false alarm.
Synthetic mismatches are not human-verified absences.

\clearpage
\section{Qualitative Cases by Selection Requirement}
\label{app:cases}

Six selected successful videos illustrate ranking, set filtering, and
query-level rejection using an earlier fixed-threshold RL variant at step 200
with 150 online head updates. These historical mechanism examples are distinct
from the final model used in every quantitative table.
Complete-split analyses of that final model are in Appendix~\ref{app:fixed-pool}.
Two additional cases from the final model below examine where selection
cannot recover an annotated occurrence.

\begin{figure}[H]
  \centering
  \includegraphics[width=\linewidth]{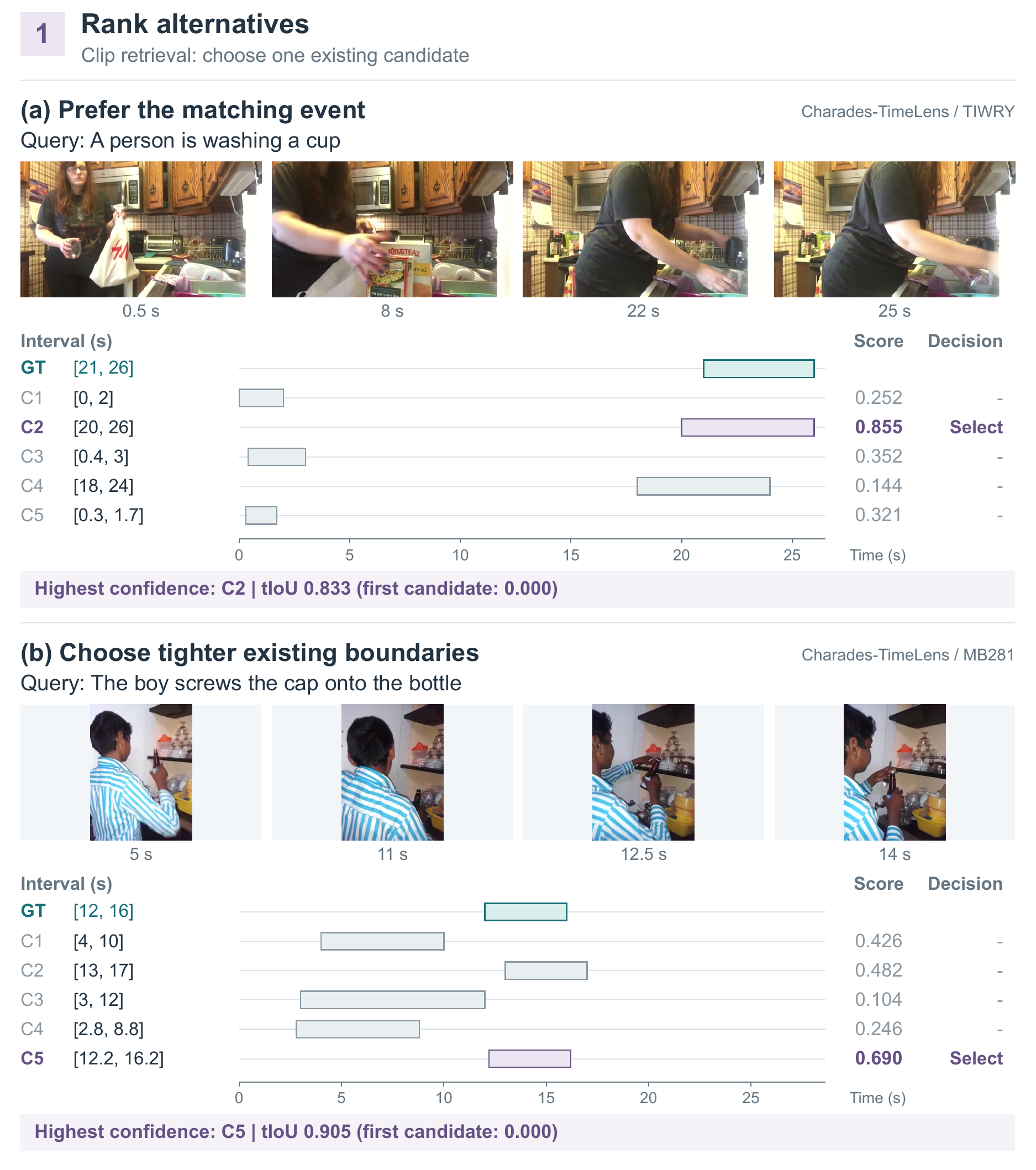}
  \caption{\textbf{Ranking: a workflow needs one useful clip from several proposals.}
  \textbf{(a)} Confidence selects the later washing event instead of the first
  generated interval. \textbf{(b)} For replacing the bottle cap, it selects
  an existing boundary alternative with tIoU 0.905; the earlier $C_2$ has
  tIoU 0.600. This is selection among decoded intervals, with no boundary
  refinement. All raw candidates are shown in generation order; the
  highest score supplies the single answer, without NMS or thresholding.
  Across the case figures, teal is GT, purple is selected, and gray is
  unselected. Scores are rounded only for display; all times are clip-relative seconds.}
  \label{fig:confidence-test-cases}
\end{figure}

\clearpage

\paragraph{Geometric overlap does not determine acceptance.}
In query 132, the dropped $[11,23]$ interval overlaps the earlier $[18,28]$
proposal at tIoU 0.294, so it survives the fixed NMS rule. Its maximum GT
IoU is only 0.400, and its confidence is 0.107. Query 114 makes the cardinality
distinction explicit: thresholding keeps four separate occurrences, rather
than imposing a top-1 or fixed top-$k$ budget. These behaviors are useful
when a review queue should preserve multiple supported events while reducing
unnecessary clips.

\clearpage
\begin{figure}[H]
  \centering
  \includegraphics[width=\linewidth]{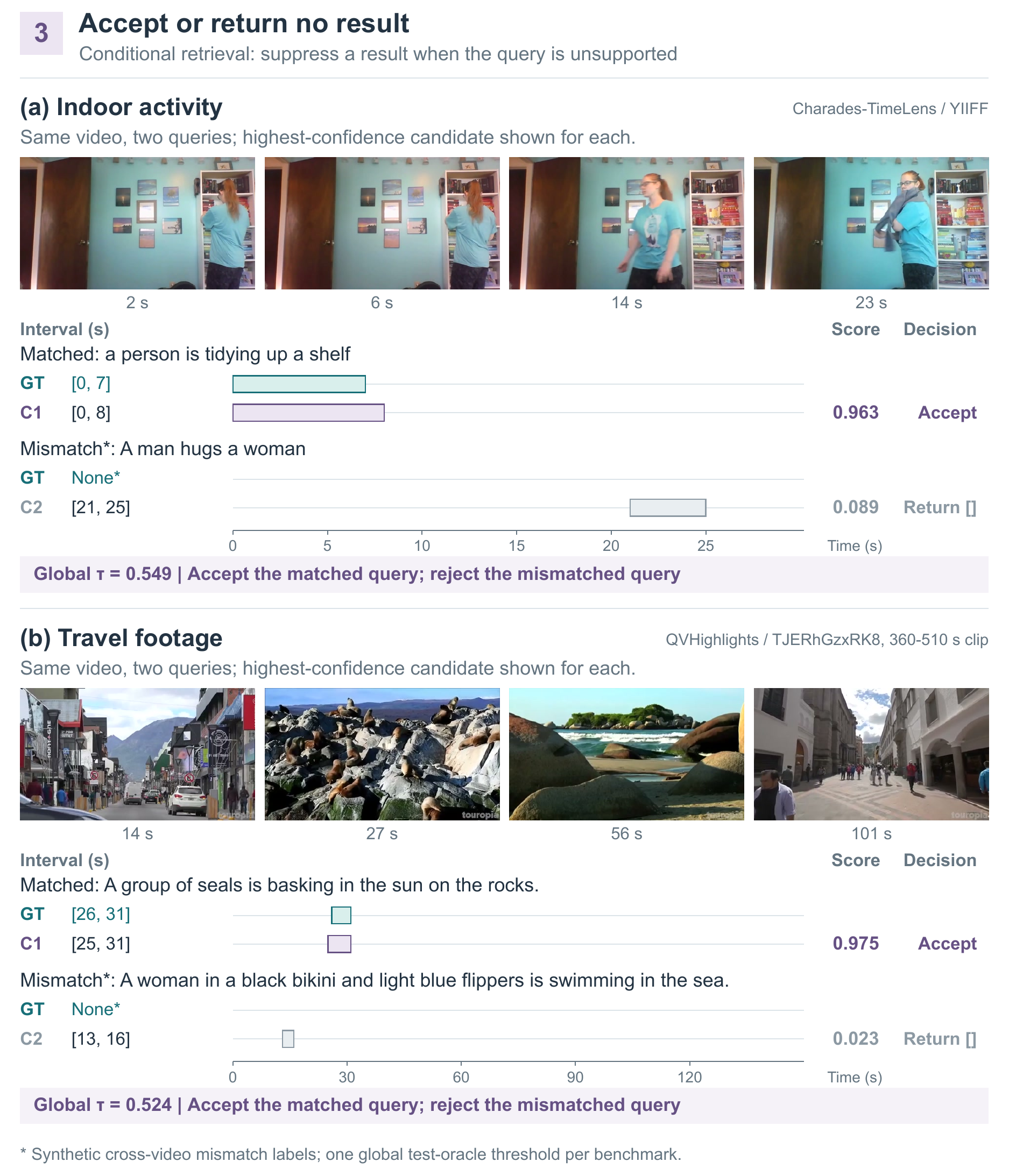}
  \caption{\textbf{Query-level rejection: a retrieval interface may need to return no result.}
  Each video is paired with its original matched query and one synthetic
  cross-video mismatch from the accepted mixed-query evaluation.
  Only the highest-confidence raw candidate for each query is displayed;
  the generator produced nonempty candidates for all four queries.
  For this raw-pool diagnostic, the benchmark-wide test-oracle thresholds are 0.549 for Charades and
  0.524 for QVHighlights, with no per-case tuning and no NMS.
  Both matched queries are accepted, while both mismatches return $[]$.
  An asterisk denotes a synthetic unmatched label, not an official
  exhaustively verified negative annotation.}
  \label{fig:confidence-case-rejection}
\end{figure}

\paragraph{Ranking and rejection answer different questions.}
A maximum always exists in a nonempty candidate pool, even when the query
is unsupported. On the indoor video, the best scores are 0.963 for tidying
and 0.089 for the mismatched interaction; on the travel video they are 0.975
for the animals on the rocks and 0.023 for the swimming query.
The threshold decides whether to return the best candidate at all.
Thus confidence can support conditional search or a gate before an automatic
action, in addition to ordering clips for human inspection.

\clearpage
\subsection{Limits of Selection on a Fixed Candidate Pool}
\label{app:selection-limits}

\begin{figure}[H]
  \centering
  \includegraphics[width=\linewidth]{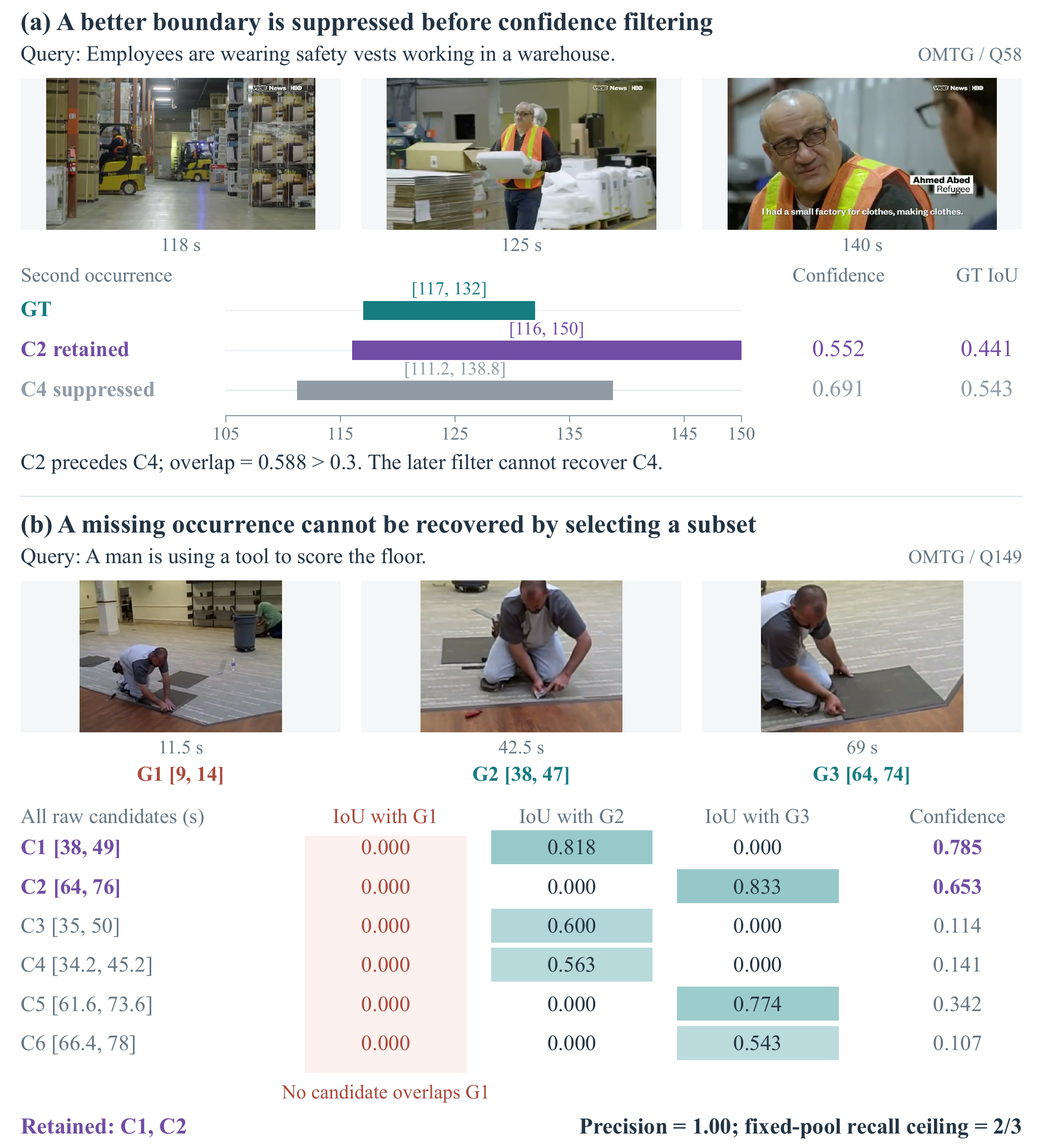}
  \caption{\textbf{Two routes to a missed occurrence.}
  Selected OMTG-Bench cases from the final L200 model with 150 head updates.
  \textbf{(a)} A close-up of the second annotation in Q58: generation-order
  NMS suppresses C4 despite its higher confidence and GT IoU.
  \textbf{(b)} All six raw candidates in Q149 miss G1; purple rows are
  retained by NMS and the model's global F1 test-oracle threshold.
  Matrix entries are original-coordinate tIoUs; matching uses tIoU $\geq0.5$.
  Times are clip-relative seconds. Frames provide scene context, and candidate
  IDs preserve generation order.}
  \label{fig:confidence-selection-limits}
\end{figure}

\paragraph{Suppression and missing proposals impose different limits.}
In Q58, a usable boundary exists but is removed before confidence filtering;
in Q149, no decoded candidate overlaps the first occurrence, so even perfect
subset selection cannot exceed $2/3$ recall. These cases separate representative
selection within overlapping proposals from proposal coverage. They motivate
diagnosing both stages, without establishing that a different NMS policy
improves overall performance.

\end{document}